\documentclass[sigconf]{acmart}

\usepackage{graphicx}
\usepackage{subcaption}
\usepackage{booktabs}
\usepackage{multirow}
\usepackage{bm}
\usepackage{tabularx}
\usepackage{amsmath}
\usepackage{array}
\usepackage{ulem}
\usepackage{xcolor}
\usepackage{threeparttable}

\AtBeginDocument{
  \providecommand\BibTeX{{
    \normalfont B\kern-0.5em{\scshape i\kern-0.25em b}\kern-0.8em\TeX}}}

\copyrightyear{2024}
\acmYear{2024}
\setcopyright{rightsretained}
\acmConference[CIKM '24]{Proceedings of the 33rd ACM International Conference on Information and Knowledge Management}{October 21--25, 2024}{Boise, ID, USA}
\acmBooktitle{Proceedings of the 33rd ACM International Conference on Information and Knowledge Management (CIKM '24), October 21--25, 2024, Boise, ID, USA}
\acmDOI{10.1145/3627673.3679757}
\acmISBN{979-8-4007-0436-9/24/10}

\begin{document}
\title{Scalable Transformer for High Dimensional Multivariate Time Series Forecasting}

\author{Xin Zhou}
\affiliation{
  \institution{Monash University}
  \city{Melbourne}
  \state{Victoria}
  \country{Australia}
}
\email{xin.zhou@monash.edu}

\author{Weiqing Wang}
\authornote{Corresponding author.}
\affiliation{
  \institution{Monash University}
  \city{Melbourne}
  \state{Victoria}
  \country{Australia}}
\email{teresa.wang@monash.edu}

\author{Wray Buntine}
\affiliation{
  \institution{VinUniversity}
  \city{Hanoi}
  \country{Vietnam}}
\email{wray.b@vinuni.edu.vn}

\author{Shilin Qu}
\affiliation{
  \institution{Monash University}
  \city{Melbourne}
  \state{Victoria}
  \country{Australia}}
\email{shilin.qu@monash.edu}

\author{Abishek Sriramulu}
\affiliation{
  \institution{Tymestack}
  \city{Melbourne}
  \state{Victoria}
  \country{Australia}}
\email{abishek.sriramulu1@monash.edu}

\author{Weicong Tan}
\affiliation{
  \institution{Monash University}
  \city{Melbourne}
  \state{Victoria}
  \country{Australia}}
\email{weicong.tan@monash.edu}

\author{Christoph Bergmeir}
\authornote{Also serve for Monash University, Melbourne, Australia.}
\affiliation{
  \institution{University of Granada}
  \city{Granada}
  \country{Spain}}
\email{bergmeir@ugr.es}

\begin{abstract}
Deep models for Multivariate Time Series (MTS) forecasting have recently demonstrated significant success. Channel-dependent models capture complex dependencies that channel-independent models cannot capture. However, the number of channels in real-world applications outpaces the capabilities of existing channel-dependent models, and contrary to common expectations, some models underperform the channel-independent models in handling high-dimensional data, which raises questions about the performance of channel-dependent models.
To address this, our study first investigates the reasons behind the suboptimal performance of these channel-dependent models on high-dimensional MTS data. Our analysis reveals that two primary issues lie in the introduced noise from unrelated series that increases the difficulty of capturing the crucial inter-channel dependencies, and challenges in training strategies due to high-dimensional data. To address these issues, we propose STHD, the \textbf{S}calable \textbf{T}ransformer for \textbf{H}igh-\textbf{D}imensional Multivariate Time Series Forecasting. STHD has three components: a) Relation Matrix Sparsity that limits the noise introduced and alleviates the memory issue; b) ReIndex applied as a training strategy to enable a more flexible batch size setting and increase the diversity of training data; and  c) Transformer that handles 2-D inputs and captures channel dependencies. These components jointly enable STHD to manage the high-dimensional MTS while maintaining computational feasibility. Furthermore, experimental results show STHD's considerable improvement on three high-dimensional datasets: Crime-Chicago, Wiki-People, and Traffic. The source code and dataset are publicly available \footnote{https://github.com/xinzzzhou/ScalableTransformer4HighDimensionMTSF.git}.
\end{abstract}


\maketitle

\section{Introduction}
Multivariate Time Series (MTS) forecasting involves predicting future values for multiple variables simultaneously based on historical time series data. This technique shows promise and applicability across various domains
and is gaining attention due to its wide applications in forecasting
natural disasters~\cite{gerard_wildfirespreadts_nodate,andrea_nascetti_biomassters_nodate,10.1145/3583780.3614966,10.1145/3583780.3614857,9785216,10.1145/3588951,10.1145/3589270,10.1145/3583780.3615069,10415669}, government management~\cite{wu_connecting_2020,zhang_crossformer_2023,liu2024itransformer,10.1145/3583780.3615136,10.1145/3583780.3614962,10.1145/3583780.3614944}, monitoring human health ~\cite{tan_monash_2020,lan2024better,wang2023contrast,Zhang2023Warpformer,10.1145/3583780.3614840,10.1145/3580305.3599543,10027783}, 
and decision-making~\cite{10.1145/3534678.3539416,10.1145/3583780.3614671,10.1145/3511808.3557470,10.1145/3511808.3557089,10.14778/3611479.3611532,10.1145/3511808.3557294,10.1145/3511808.3557091,10.1145/3583780.3615478,10.1145/3583780.3615473,10.1145/3583780.3614751,10.1145/3583780.3614918}, etc.
MTS differs from univariate time series by consisting of multiple time-dependent variates, each contributing to the dynamics of the dataset.
Recently, the increasing amount of large-scale MTS data has led to two significant developments:
the extension of time dimensions (long-time dimension) and a substantial increase in the number of channels (high-dimensional)~\footnote{\textbf{`channel', `dimension', and `variate' are interchangeable in this work.}}, each bringing new challenges to statistical forecasting methods.
Statistical forecasting methods~\cite{BoxJenkins1970ARIMA,Lutkepohl1985VARIMA,Hyndman2008ETS}, are often designed according to the characteristics of the data, which usually will become very complex for large-scale MTS data.

Deep Neural Network (DNN) models have shown great success in many areas, because of their ability to model non-linear transformations, versatility, and scalability.
Given the critical assumption for time series data that future values depend on historical values, mainstream research mainly focuses on the long-time data~\cite{Zeng2022AreTE,xu_tensorized_2020,nie_time_2023} and most existing works simply model high-dimensional data
with channel (or dimension)-independent
approaches~\cite{mohsen_forecasting_2023}.
This focus has led to significant advances in understanding and modeling temporal dynamics over long-time periods.
Transformer-based methods emerge as a promising technique for enhancing temporal analysis, offering significant performance in data mining and predictive
It is a mature DNN model that is now popular for time series ~\cite{ekambaram_tsmixer_2023,zhou_informer_2021,nie_time_2023,zhou_fedformer_2022}. 
PatchTST~\cite{nie_time_2023}, iTransformer~\cite{liu2024itransformer}, and Crossformer~\cite{zhang_crossformer_2023}) are leading transformer-based methods that show great performance in MTS forecasting. They represent two categories of MTS forecasting technologies: channel-independent and channel-dependent models. These terms refer to how various dimensions in MTS data are treated relative to each other during the training process. However, same with the other mainstream DNN models, they focus more on long-time forecasting instead of high-dimensional MTS forecasting. 
In this work, we focus on using modern transformer architectures to address the gap for high-dimensional MTS forecasting. 
Our motivation is underscored by existing works~\cite{zhao_multi-type_2022,wang_hagen_2022,zhang_crossformer_2023}, which emphasize the importance of learning relations among variables for prediction (i.e., channel-dependent modeling).

Applying channel-dependent transformers to high-dimensional MTS forecasting presents specific challenges, including learning complex segment relationships from across all channels, and coping with increased memory and computational complexity.
One of the primary challenges we observed and reported in this paper is that some existing channel-dependent models underperform channel-independent models on high-dimensional data, contrary to common expectations that accounting for dimension relationships enhances forecasting accuracy. Specifically, on three datasets: Crime-Chicago, Wiki-People, and Traffic, one state-of-the-art channel-dependent transformer, Crossformer, is on average 19\% less accurate than the state-of-the-art channel-independent transformer, PatchTST. 
Another challenge we observe is the computational overhead in modeling correlations among all the segments in a multitude of interacting channels. 
Existing work can only model the time relations within individual series~\cite{nie_time_2023}, or for channel relations within individual segment~\cite{zhang_crossformer_2023}. 
A na\"ive solution is to learn the attention map in the shape of $(M \times S) \times (M \times S)$, where $S$ is the number of subseries split from each series, and $M$ is the total number of series.
However, learning a huge attention map directly imposes a significant computational burden. 
Furthermore, the scale of high dimensions strains memory and processing power. It limits existing channel-dependent models from being designed with large values for parameters such as batch size and embedding size.
Take the Crossformer as an example. The batch sampling size is $b \times M$, where $b$ is the batch size. For the high-dimensional data, $M$ is large, increasing the memory and processing power usage, making it $M$ times larger than in usual cases. To ensure the algorithm runs successfully, a compromise solution is to limit the tuning range of parameters (e.g., setting smaller batch sizes), but this leads to the sacrifice in model performance.

Our research aims to bridge the above gaps, and at the same time, to retain the performance as a channel-dependent model. Therefore, we propose a novel approach to scalably manage the channels, their comprehensive dependencies, and parameter settings along with the extended time dimensions inherent in high-dimensional MTS. 
To the best of our knowledge, 
we are the first to use the modern Transformer architecture for high-dimensional MTS forecasting, 
with the ability to explore information from dependencies between channels.

Our key contributions are listed as follows:
First, we conduct a detailed analysis to understand the reasons behind the suboptimal performance of channel-dependent models on high-dimensional MTS data, identifying the main issue is the introduced noise by possibly many unrelated channels. 
Second, we propose STHD, the \textbf{S}calable \textbf{T}ransformer for \textbf{H}igh-\textbf{D}imensional Multivariate Time Series Forecasting. 
    To ensure efficiency, STHD designs a 2-D transformer with a sparsed attention map to be learned enabled by  DeepGraph~\cite{traxl-2016-deep} which sparsifies the relations between a large number of channels~\cite{sriramulu2023adaptive}. Apart from the improved learning efficiency, enforcing sparsity in the 2-D transformer also reduces introduced noise, and reduces memory and processing power consumption.
Third, channel-dependent MTS models need large numbers of parameters, causing memory issues. Our proposed ``ReIndex'' strategy
optimizes computational resources and increases the diversity of shuffled training samples, ultimately aiding STHD in better generalization.
Finally, STHD presents a significant improvement in high-dimensional MTS forecasting on three real-world high-dimensional datasets: Crime-Chicago, Wiki-People, and Traffic, which have higher dimensions compared with the datasets used in the existing works. In our experimental evaluations, STHD achieved an average 8.91\% performance improvement on the three datasets.

\section{Related Work}
\label{sec:related-work}
\subsection{Channel-independent Model}
A channel-independent model treats each channel as an individual series, to forecast with just the given series. 
For the data whose channels may not have significant dependencies, channel-independent modeling simplifies the process.
Statistical methods in this field are mostly global models or local univariate models, adapting to the features of data. 
One of the most prevalent approaches is Auto-Regressive Integrated Moving Average(ARIMA)~\cite{BoxJenkins1970ARIMA}, which incorporates autoregression, differencing, and moving averages, providing a detailed understanding of time series dynamics.
Another fundamental statistical method is Exponential Smoothing (ETS)~\cite{Hyndman2008ETS} applies weighting factors that decrease exponentially over time. ETS is widely used for smoothing time series data to identify trends and patterns.
However, the computation speed and memory of the above methods limit their scalability for long, and high-dimensional data analysis.\\

DNNs make a significant advance in handling sequential time series data.
Recurrent Neural Networks (RNNs), particularly Long Short-Term Memory (LSTM), represent channel-independent parad-igms~\cite{xu_tensorized_2020,salinas2019high}.
LSTM
learns long-term dependencies, making them suitable for modeling complex temporal dynamics in MTS forecasting. Their ability to remember information over extended periods and to handle large-scale data address the limitation of statistical methods.
Recent advances in deep learning for forecasting, focus on Convolutional Neural Networks (CNNs)~\cite{10.1145/3583780.3614876,10223408,10.1145/3583780.3615164,10.1145/3511808.3557135} and attention-based models~\cite{10.1145/3583780.3614851,10.1145/3583780.3615253,10.1145/3511808.3557386,10.1145/3511808.3557077,10.1145/3511808.3557705,9807714,chang2024timedrl,10184599,10.1145/3580305.3599549,10184766,10184658}.
DeepGLO~\cite{sen2019think}, designed for univariate time series forecasting, can learn both global and local views of all series through CNNs.
Time Series with Transformer (TST)~\cite{zerveas_transformer-based_2021} is the first work that applies transformer to MTS forecasting.
Informer~\cite{zhou_informer_2021} focuses on efficient transformer forecasting.
FEDformer focuses on the time-frequency analysis of individual series.
PatchTST~\cite{nie_time_2023} divides series into different patches (segments) and through the Transformer framework learns dynamic time relationships.
TSMixer~\cite{ekambaram_tsmixer_2023} replaces the attention mechanism with linear layers to improve computation efficiency.
TimesNet~\cite{wu2023timesnet} splits time series into different frequency domains to capture comprehensive temporal dynamics.
LLM-based models~\cite{Zhou2023OneFA,jin2023time,jia_gpt4mts_2024} first tokenize series, then extract information in pre-trained language models, which is time-consuming for fine-tuning than transformer's training.
The above methods, especially transformers, can handle long sequences more effectively and provide more flexibility in capturing complex temporal relations, so we intend to design a transformer-based model to solve high-dimensional MTS forecasting.

\subsection{Channel-dependent Model}
The channel-dependent model recognizes and leverages the dependencies among channels, that is to forecast with both target series and auxiliary series.
Recognizing these dependencies can lead to more accurate and explainable forecasts.
The fundamental statistical method is Vector Auto-Regression (VAR)~\cite{Sims1980VAR}, which captures linear dependencies among multiple time series. Despite a simple and effective framework for MTS analysis, VAR might not perform well with non-linear data.
Besides, some statistical methods are used for channel dependency analysis in different aspects~\cite{sriramulu2023adaptive}: Correlation Analysis, Granger Causality, Mutual Information, Maximum Likelihood Estimation, Naive Transfer Entropy, Optimal Causation Entropy, 
and Thouless-Anderson-Palmer. 
The above methods are hard to scale for high-dimensional data due to slow computation.

DNNs can be channel-dependent in linear and non-linear ways.
Representative linear methods are Linear, Normalization Linear (NLinear), and Decomposed Linear (DLinear)~\cite{Zeng2022AreTE}. They are linear model implemented in the Deep-learning framework.
DLinear applies a linear model on the decomposed elements in different frequency domains of time series. With this, DLinear can precisely capture the seasonality, trend, and residual components.
DeepVAR~\cite{salinas2019high} is an RNN-based time series model with a Gaussian copula process output model.
TST~\cite{zerveas_transformer-based_2021} is a pre-trained model for time series representation learning. 
STEP~\cite{shao_pre-training_2022} learns both the spatial and temporal patterns based on TST.
Crossformer~\cite{zhang_crossformer_2023} develops a transformer-based model to learn both relations cross time and cross dimension, separately.
Graph Neural Networks (GNNs) are wildly used in forecasting~\cite{wu_connecting_2020,10.1145/3511808.3557702,10.1145/3511808.3557638,ma2023learning,chen2023multiscale,10069881,10.14778/3489496.3489503,10.1145/3580305.3599444,10.1145/3511808.3557540,10.1145/3583780.3615066,10.1145/3583780.3615200,duan2023localised}.
Ye et al.~\cite{10.1145/3534678.3539274} model time series with a hierarchical graph structure to capture the scale-specific correlations among series.
FourierGNN~\cite{yi2023fouriergnn} define time series data as a hypervariable graph structure and a Fourier learning method.
The evolving nature of time series necessitates that the above models be adapted to dynamic patterns and incorporate diverse data characteristics.
There is a growing trend towards hybrid models that blend statistical and machine learning techniques. ADLNNs~\cite{sriramulu2023adaptive} first apply statistical methods to the sparse correlation matrix, then apply GNNs to assist forecasting.
They can capture the dynamic co-evolutions and use the information from other channels to improve the forecast of the target one.
The main challenge lies in effectively modeling the complex and possible non-linear dependencies. This often requires complex model design, thus leading to high computational memory requirements.

\section{Initial Analysis}
\begin{figure}[htp]
    \begin{subfigure}{.115\textwidth}
        \centering
        \includegraphics[width=\linewidth]{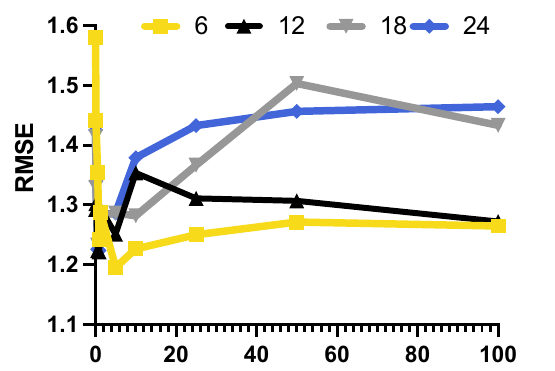}
        \caption{Crossformer}
        \label{fig:pretest_cross_rmse}
    \end{subfigure}
    \begin{subfigure}{.115\textwidth}
        \centering
        \includegraphics[width=\linewidth]{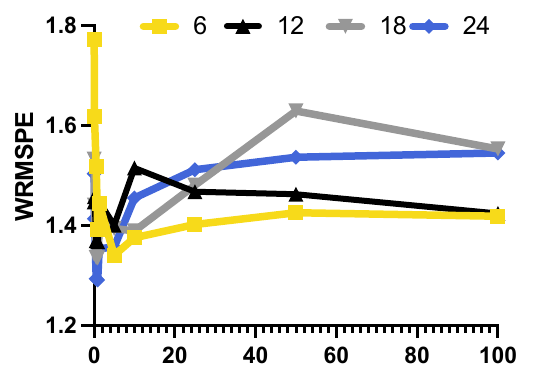}
        \caption{Crossformer}
        \label{fig:pretest_cross_wrmspe}
    \end{subfigure}
    \begin{subfigure}{.115\textwidth}
        \centering
        \includegraphics[width=\linewidth]{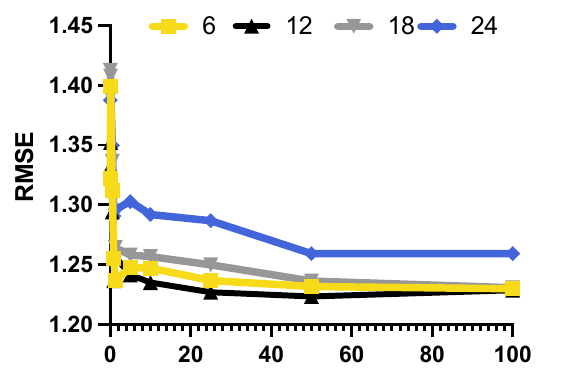}
        \caption{iTransfor}
        \label{fig:pretest_itrans_rmse}
    \end{subfigure}
    \begin{subfigure}{.115\textwidth}
        \centering
        \includegraphics[width=\linewidth]{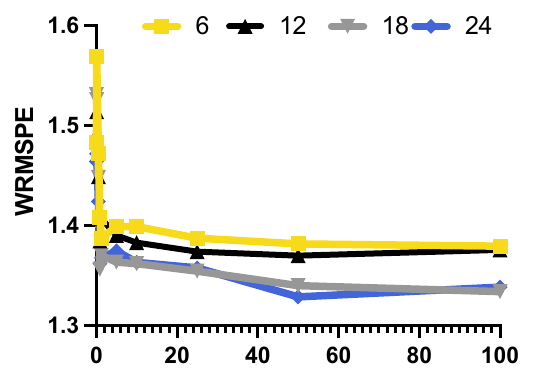}
        \caption{iTransfor}
        \label{fig:pretest_itrans_wrmspe}
    \end{subfigure}
    \caption{Result of using different proportions of training data on Crossformer and iTransformer. The x-axis describes proportions from 0\% to 100\%. The y-axis in (a) and (c) represents RMSE on Crossformer and iTransformer (iTransfor for short), and the y-axis in (b) and (d) represents WRMSPE on Crossformer and iTransformer. Lines in yellow, black, grey, and blue represent the horizon of \{6,12,18,24\} months.}
    \Description{}
    \label{fig:pretest}
\end{figure}

In this section, we conduct exploratory experiments and data analysis on a high-dimensional dataset: Crime-Chicago, a precursor to the model development phase.
We employ Crossformer~\cite{zhang_crossformer_2023} and iTransformer~\cite{liu2024itransformer}, state-of-the-art channel-dependent transformer models, as our pre-test models.
We start with a randomly selected series from the dataset as the target series. Then, we use the last 20\% as the test sub-series. Subsequently, we sample different proportions of top $K$ correlated series with target series as training data, where $K \in \{0, 1, 5, 10, 15, 57, 114, 288, 576, 1154\}$, representing the proportions in \{0\%, 0.1\%, 0.5\%, 0.9\%, 1.3\%, 5\%, 10\%, 25\%, 50\%, 100\%\} respectively.
Figure~\ref{fig:pretest} shows the performance of training with the above data for Crossformer and iTransformer (iTransfor for short in Figure~\ref{fig:pretest}).

\begin{figure}[htp]
    \centering
    \begin{subfigure}{.18\textwidth}
        \centering
        \includegraphics[width=\linewidth]{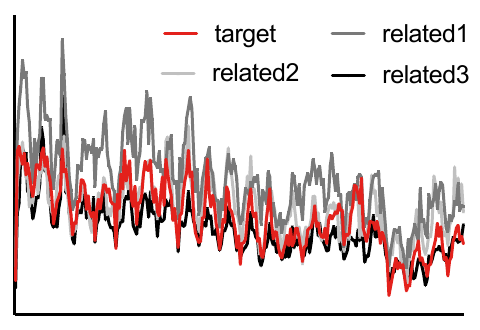}
        \caption{target series with top 3 related series}
        \label{fig:visual_rela}
    \end{subfigure}
    \hspace{0.5cm}
    \begin{subfigure}{.18\textwidth}
        \centering
        \includegraphics[width=\linewidth]{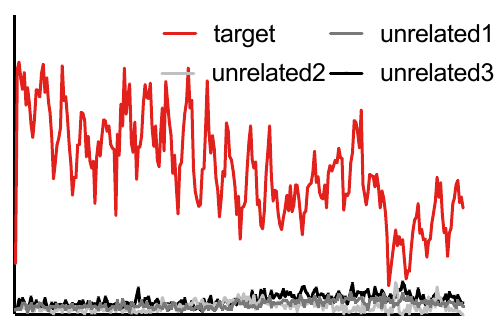}
        \caption{target series with top 3 unrelated series}
        \label{fig:visual_unrela}
    \end{subfigure}
    \caption{Visualisation of the target series in \textcolor{red}{red}, and its related series/non-related series in \textcolor{gray}{gray} and \textcolor{black}{black}.}
    \Description{}
    \label{fig:eda_crime}
\end{figure}

Please note that, both metrics in Figure~\ref{fig:pretest} are metrics of errors which means that they both are the less the better. 
From Figure~\ref{fig:pretest_cross_rmse}-\ref{fig:pretest_cross_wrmspe}, we find a sharp metric downtrend at the very beginning and then was followed by an obvious rise.
From Figure~\ref{fig:pretest_itrans_rmse}-\ref{fig:pretest_itrans_wrmspe}, we find when the proportion equals 50\%, the performance is the best. This shows that for both channel-dependent models, for the target series, utilizing some series achieves the best performance compared with utilizing all series or none at all. We guess the reason is that
utilizing 100\% of the series as training data across channel-dependent models, which have full connections to all series, can potentially introduce noise stemming from unrelated connections. 
That means, a proper number of related series, may introduce useful patterns to forecasting. To confirm this, we perform the data analysis in Figure~\ref{fig:eda_crime}. 

Figure~\ref{fig:visual_rela} and Figure~\ref{fig:visual_unrela} compare the visualization of the same target series (in red) with its related series Figure~\ref{fig:visual_rela} and unrelated series Figure~\ref{fig:visual_unrela}.  We can observe that related series show similar trends with target series, while unrelated series show different and irregular trends.
\begin{figure*}[h]
    \centering
    \includegraphics[width=0.68\textwidth]{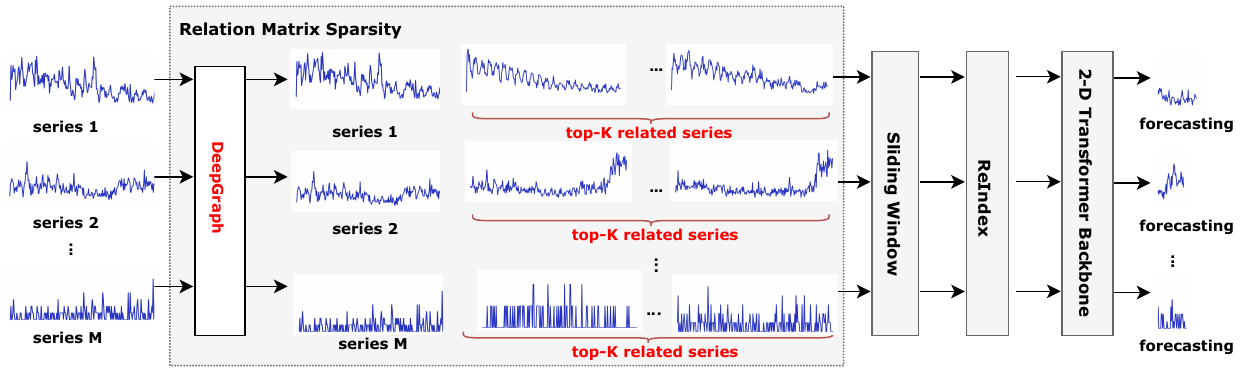} 
    \caption{Overview of STHD, correlations of Multivariate Time Series data from different aspects are computed and then sparsed. That is, each series reserves the top $K$ auxiliary series. Then, the target series together with $K$ auxiliary series are input to the 2-D Transformer Backbone, which can extract the representation of 2-D input.}
    \Description{}
    \label{fig:framework}
\end{figure*}
Overall, in high-dimensional data, introducing unrelated patterns, and unrelated series can lead to suboptimal performance.

\section{Scalable Transformer STHD }
\paragraph{Problem Formulation}
We are solving Multivariate Time Series Forecasting with a channel-dependent model for high-dimensional data.
Following the common practice of existing works~\cite{nie_time_2023,zhang_crossformer_2023}, we use the sliding window to split each series into $S$ subseries, with each containing an input and a horizon, when training, the loss computes a little overlap of series, which is similar to the common practice in NLP. 
Given the historical observation set of time series $\mathcal{X} = \{X_1, X_2, ..., X_M\} \in \mathbb{N}^{M \times T \times (1+K)}$,  which consists of $M$ series with 1 target series, and $K$ related series for each, all of them are in $T$ time points.  For example, in the Wiki-People prediction dataset, the series $\mathcal{X}$ is the traffic of each article, and $K$ articles that show a similar temporal pattern.
The output is the future $\tau$ time points' observations (Horizon) $\mathcal{Y} = \{Y_1, Y_2, ..., Y_M\} \in \mathbb{N}^{M \times \tau}$ for $M$ series.\\
After the sliding window process, the number of windows $S = \left\lfloor \frac{T-1}{L} \right\rfloor + 2$, where $L$ is the window length, the stride is set to 1. Finally, the input $\mathcal{X}^{'} \in \mathbb{N}^{M \times S \times L \times (1+K)}$, the output $\mathcal{Y}^{'} \in \mathbb{N}^{M \times S \times \tau}$.

\paragraph{Overview of STHD}
The supervised learning pipeline is shown in Figure~\ref{fig:framework}. 
First, STHD starts with the Relation Matrix Sparsity module, which refines the input data by focusing on the most relevant inter-channel relationships. According to relation matrix $\Gamma \in \mathbb{R}^{M \times M}$, STHD gets $K$ related series with $1$ target series as input. This not only enhances the model's ability to capture channel dependencies but also alleviates the memory issue commonly encountered in handling high-dimensional data. This content is introduced in Section 4.1. 
Then, the refined data passes through the ReIndex module, which ensures efficient data processing and enhances training diversity, ultimately aiding in better model generalization.
We introduce this content in Section 4.2. 
Finally, the 2-D Transformer, introduced in Section 4.3, with its capability to handle 2-D inputs, utilizes the processed and optimized data to capture the complex dependencies between channels over time. STHD predicts future $\mathcal{Y}^{'}$. 
Collectively, these modules are optimized for computational efficiency, making STHD viable for high-dimensional MTS forecasting.

\subsection{Relation Matrix Sparsity}
Based on the assumption in Section 3 that not all inter-series relationships contribute equally to forecasting accuracy, STHD uses Pearson Correlation, an efficient method, to capture the extent of series co-variates. The impact of this component is twofold: it improves forecasting accuracy by reducing the influence of noise and irrelevant information and improves computational efficiency with DeepGraph as it allows the model to focus on the related series instead of all the series.

Referring to ADLNN~\cite{sriramulu2023adaptive}, which demonstrates the effectiveness of correlation matrix sparsity, we use correlation matrix sparsity to get $K$ related series. However, the computation is very slow, especially for high-dimensional MTS data. To facilitate efficient relation computation and manage the relationships within the dataset, we use DeepGraph~\cite{traxl-2016-deep}, a structure used for computing relations, to model high-dimensional MTS data by formulating individual series as a node and relation with another series as an edge. It enables the dynamic construction of a sparse correlation matrix by identifying and mapping significant inter-series relationships.
The fast computation of pairwise correlation matrices in the DeepGraph framework is attributed to a combination of advanced data structures, robust scalability, and parallelization. Central to its efficiency are optimized data structures, tailored to handle complex and large datasets efficiently with full control over RAM usage. This is complemented by the parallelization and integration of Cython~\footnote{\url{https://cython.org/}}, which significantly boosts performance.
To get the sparse correlation matrix $\mathbf{\Gamma}$, for each series, we keep the top $K$ most related series, and set them as auxiliary series to assist target series forecasting. We compute ranks for each column of each matrix, to get the ranking matrix $R_{\mathbf{\Gamma}}$. Through $R_{\mathbf{\Gamma}}$, we select the top $K$ series for each series.

\begin{figure*}[ht]
    \centering
    \includegraphics[width=0.65\textwidth]{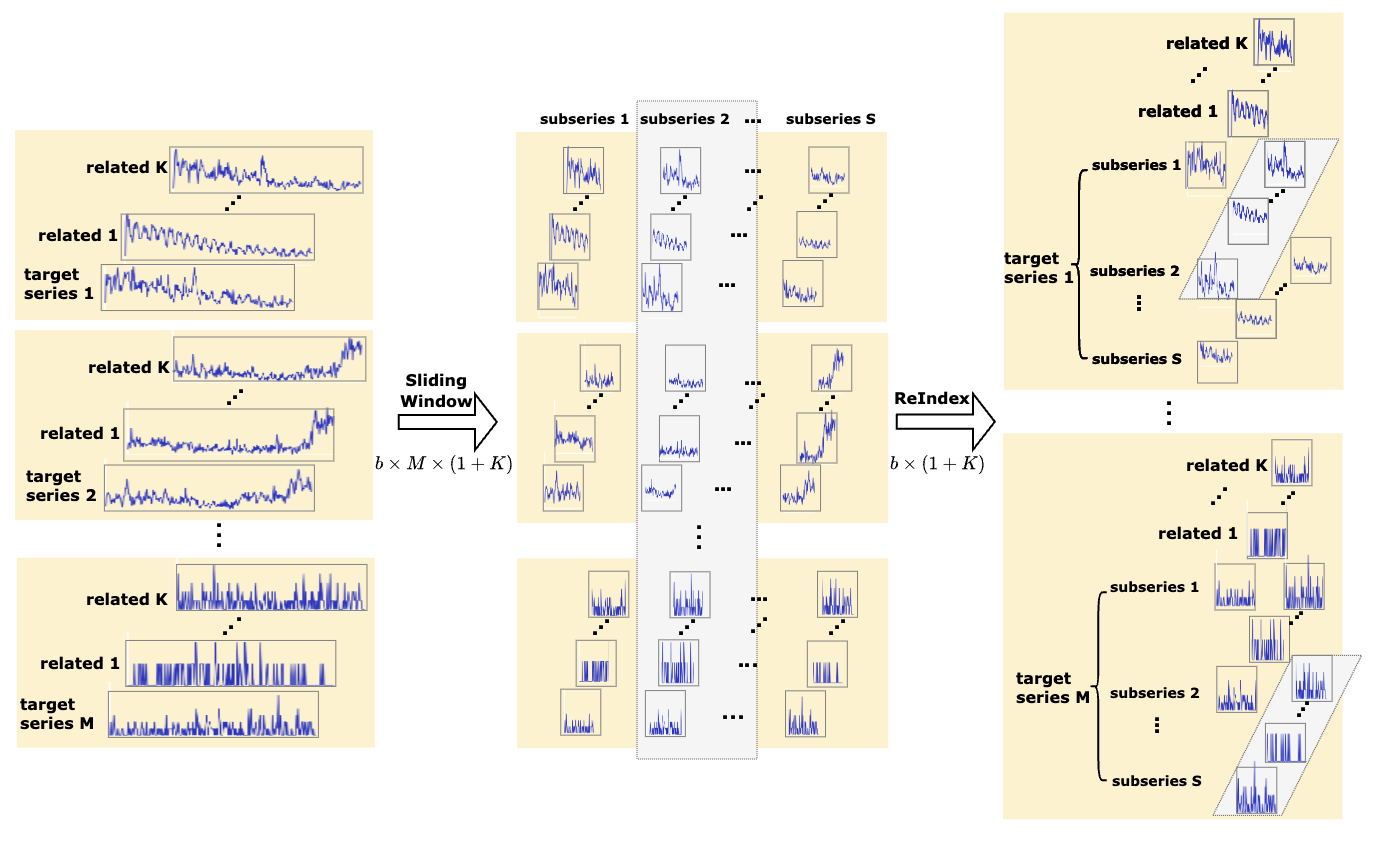}
    \caption{ReIndex process before batch sampling. Yellow square represents a series sample, including one target series, and $K$ related series. All $M \times (1+K) $ series are split to $M \times (1+K) \times S$ subseries, and each of length $L$. Existing works sample batches on the $S$ dimension. For example, the grey square in the middle of the figure denotes a batch of samples, with number of samples $b \times M \times (1+K)$, which largely increases memory usage. ReIndex reshapes the windows to $(M\times S) \times (1+K)$ and samples batches on the $M \times S$ dimension. The number of samples of each batch is $b\times (1+K)$.}
    \label{fig:reindex}
    \Description{}
\end{figure*}

\subsection{ReIndex}
The above training data $\mathcal{X}^{'} = \{\bm{x}^{(1)},\bm{x}^{(2)},...,\bm{x}^{(M \times S)}\}$ contains $M \times S$ samples in shape of $L \times (1+K)$. For simplicity, we label the $i$-th sample as $\bm{x}^{(i)} \in \mathbb{N}^{L \times (1+K)}$.
If we follow existing work to sample batches in the $S$ dimension, that is, each batch will be allocated $ b \times M \times (1+K)$ data, and each with the length of $L$, as shown in the middle of Figure~\ref{fig:reindex}, where $b$ denotes the batch size. According to existing transformer methods~\cite{zhang_crossformer_2023,nie_time_2023}, the average channel number is 200, that is $M=200$. For such a relatively small number of channels, most of the DNNs including linear, CNNs, RNNs, and transformers should not have problems. 
However, in high-dimensional datasets, transformers may incur memory issues.
In this work, to adapt transformer to high-dimension data, we design a more flexible method called ReIndex. Specifically, we sample the batch from the $M \times S $ dimensions, as shown on the right-hand side of Figure~\ref{fig:reindex}.
With ReIndex, the number of samples of each batch can be reduced from $b \times M \times (1+K)$ to $b \times (1+K)$, each with a length of $L$. This indicates that the memory usage and computational complexity of the attention map are largely decreased, so we can set more flexible batch sizes.
By setting shuffle for the training data, ReIndex improves the diversity of each training batch from specific windows in all $M$ series to any window in any series. With more diversified training examples, ReIndex is also able to improve the model's generalization ability apart from the improved memory usage and computational complexity in computing the attention map.

\subsection{2-D Transformer}
We adapt the vanilla Transformer to encode the 2-D input in the shape of $\mathbb{N}^{L \times (1+K)}$ as latent representations, where $(1+K)$ represents $1$ target series and $K$ auxiliary series.\\
\paragraph{Patches}
Patches have been proven to be effective in previous works~\cite{nie_time_2023,zhang_crossformer_2023} due to their ability to learn long and short-term temporal dependencies.
Here for each series, first we pad $l$ repeated numbers that equal the last value at the end of the original series.
Then we divide each series $\bm{x}$ of length $L$ into $P$ patches
\begin{equation}
     \bm{x}^{(i)}= \{\bm{x}^{(i)}_1,\bm{x}^{(i)}_2, ..., \bm{x}^{(i)}_P\} \in \mathbb{N}^{P \times l \times (1+K) }
\end{equation}
where $l$ is the length of a patch, $P = \left\lfloor \frac{L-l}{s} \right\rfloor + 2$ is the number of patches for each series, $s$ represents strides of getting patches.\\

\paragraph{Encoder}
First, we embed each patch $\bm{x}^{(i)}_p \in \mathbb{N}^{l \times (1+K) }$ into the $D$-dimension latent space of transformer by a trainable linear function $\mathbf{W}_{p} \in \mathbb{R}^{D \times l}$, 
and apply a sine and cosine position encoding for patches, $\mathbf{W}_{pos_{tem}}$, to distinguish the temporal order,
\begin{equation}
   \bm{z}^{(i)}_p = \mathbf{W}_{p}\bm{x}^{(i)}_p + \mathbf{W}_{pos_{tem}}[:,:1+K]
\end{equation}
Then each sample with all the patches $\bm{z}^{(i)} \in \mathbb{R}^{D \times (1+K) \times P}$.
Next, we apply positional encoding for channels, $\mathbf{W}_{pos_{cha}}$, to distinguish the target series and auxiliary series, resulting in
\begin{equation}
    \bm{z}^{(i)}_d = \bm{z}^{(i)} + \mathbf{W}_{pos_{cha}}[:,:1+K]
\end{equation}
Final output 
$O^{(i)} = \textit{MultiHeadAtn}\left(\bm{z}^{(i)}_d, \bm{z}^{(i)}_d, \bm{z}^{(i)}_d\right) \in \mathbb{R}^{(P \times (1+K) )\times D)}$.
After adding a residual connection, followed by layer normalization $\bm{z}^{(i)'}_d=\textit{LayerNorm}_1\left(\bm{z}^{(i)}_d+\ldots\right)$, $\bm{z}^{(i)'}_d$ is passed through the feedforward layer, comprising two 1-D convolutional layers $\textit{Conv}_1()$ and $\textit{Conv}_2()$.
\begin{equation}
    \bm{z}^{(i)'}_d = \textit{LayerNorm}_2\left(\bm{z}^{(i)}_d+\textit{Conv}_2\left(\textit{GeLU}\left(\textit{Conv}_1\left(\bm{z}^{(i)'}_d\right)\right)\right)\right)
\end{equation}
where \textit{GeLU}() is the Gaussian Error Linear Unit active function.

\paragraph{Attention Mechanism}
The attention mechanism is the most important element in each layer of the Transformer backbone. 
Existing works for exploring cross-time and channel relations are not enough. Channel-independent models like PatchTST use self-attention to explore time dependencies of different patches, thus ignoring the relation of channels; channel-dependent models like Crossformer use two-stage Attention, which explores relations separately.  That is, they first explore time relations, then explore channel relations. Therefore, relations for patches from different series during different periods are ignored.
In our work, we use self-attention to learn across the time and channels of target series and related series.
Here we set attention head $h \in [1,H]$. For each head, we transform $\bm{z}^{(i)}_d$ into query matrix, key matrix, and value matrix.
Then we get the attention output 
$ O^{(i)}_h \in \mathbb{R}^{(P \times (1+K) )\times D^{''}}$.
To facilitate the integration of diverse information contexts captured by the distinct attention heads,
the output of each individual attention head is concatenated into a unified representation, which is subsequently linearly transformed back to the original dimensional space $D$ of Transformer.
\begin{equation}
    O^{(i)} = \textit{Concat}(O^{(i)}_1, O^{(i)}_2, \ldots, O^{(i)}_H)\mathbf{W}_o
\end{equation}
where \textit{Concat()} denotes the concatenation operation, and $\mathbf{W}_o \in \mathbb{R}^{(H \times D^{''}) \times D}$ represents the weight matrix of the subsequent linear transformation layer mapping the concatenated output back to the model's original dimensional space.

\paragraph{Decoder}
The decoder transforms shape of $\bm{z}^{(i)'}_d$ from $\mathbb{R}^{P \times (1+K) \times D}$ to $\mathbb{R}^{(1+K) \times D}$ that is suitable for the forecasting task. Specifically, the decoder first flattens the $P$ dimension of $\bm{z}^{(i)'}_d$, then applies a linear projection to it. Finally, we get the representation of (1+K) series. STHD gets the first vector as the representation of the target series, and then transforms it to $\hat{y^{(i)}} \in \mathbb{R}^{\tau}$ that equals the horizon.
\paragraph{Loss Function} 
Following common practice~\cite{nie_time_2023,zhang_crossformer_2023}, we use Mean Squared Error (MSE) and Adam optimizer to minimize the divergence of forecasting values and ground truth values. The loss in each series is gathered and averaged to get the overall loss
$ L = \frac{1}{M\times S} \sum_{i=1}^{M\times S} (y^{(i)} - \hat{y^{(i)}})^2$,
where $y^{(i)} $ is the true value for the $i_{th}$ sample, $\hat{y^{(i)}}$ is the predicted value.

\section{Experiments}
In this section, we report our extensive experiments on real-world high-dimensional MTS datasets to answer five questions: (1) How is the performance of our STHD approach in comparison with other counterparts for high-dimensional MTS forecasting tasks (for both effectiveness and memory issues)? (2) What is the impact of using related series? (3) How effective is the efficiency of DeepGraph? (4) How does the hyperparameter $K$ influence the final performance? (5) How does ReIndex contribute to the forecasting performance? These questions can be matched to Section 5.3, Section 5.4, Section 5.5, Section 5.6, and Section 5.7 respectively.

\subsection{Datasets}
Three real-world high-dimensional MTS datasets are used for evaluation, namely Crime-Chicago, Wiki-People, and Traffic. 
Crime-Chicago and Wiki-People are larger in channel dimensions, the length is not very long. Traffic has relatively small channel dimensions, while it is longer than the other two datasets. Details are in Table~\ref{tab:dataset}.

\begin{table}[h]
    \centering
    \renewcommand{\arraystretch}{0.6}
    \caption{Details of multivariate datasets.}
    \label{tab:dataset}
    \begin{tabular}{c|c|c|c} \hline
                & Crime-Chicago & Wiki-People & Traffic\\ \hline
     channels   & 1,155 & 6,107  & 862\\
     length     & 260    &  550 & 17,545\\
     frequency  & monthly & daily  & hourly\\ \hline
    \end{tabular} 
\end{table}

\begin{table*}[h]
    \begin{threeparttable}
    \caption{Experimental results comparing with baselines. The horizon length values are from Section~\ref{ssct-settings}. The best results are in \textbf{bold} and the second best are \underline{underlined}. To save space, NTransformer is the abbreviation of Non-stationary Transformer.}
    \label{tab:main_results}
    \centering 
    \small
    \renewcommand{\arraystretch}{0.6}
    \begin{tabular}{p{1.655cm}||>{\centering\arraybackslash}p{0.57cm}|p{0.57cm}|>{\centering\arraybackslash}p{0.57cm}|p{0.57cm}|p{0.57cm}|p{0.57cm}|p{0.57cm}|p{0.57cm}|p{0.57cm}|p{0.57cm}|p{0.57cm}|p{0.57cm}|p{0.57cm}|p{0.57cm}|p{0.57cm}|p{0.57cm}}\hline
       {Dataset}    & \multicolumn{15}{c}{Crime-Chicago}\\ \hline
       & \multicolumn{4}{c|}{RMSE} & \multicolumn{4}{c|}{WRMSPE} & \multicolumn{4}{c|}{MAE} & \multicolumn{4}{c}{WAPE}\\ \hline
       Horizon      & \ 6 & \ \ 12 & \ 18 & \ \ 24 & \ \ \ 6 & \ \ 12 & \ \ 18 & \ \ 24 & \ \ 6 & \ \ 12 & \ 18 & \ 24  & \ \  6 & \ \ 12 & \ 18 & \ \ 24\\ \hline
       Naive        & 1.334 & 1.340 & 1.440 & 1.371 &  1.088 & 1.081 & 1.157 & 1.104 & 0.827 &0.816&  0.901& 0.865 & 0.674 & 0.659 & 0.724 & 0.697\\
       Linear & 2.006 & 1.794& 1.729 & 1.685 & 1.635 & 1.448 &1.389& 1.357& 1.586 & 1.358 & 1.303 &1.263 &1.293 & 1.096& 1.047& 1.017\\
       DLinear & 1.033 & 1.121 & 1.133 & 1.139 & 0.842 & 0.905 & 0.910 & 0.917 & 0.655 & 0.718 & 0.719 & 0.725 & 0.534 & 0.579 & 0.578 & 0.584\\
       Transformer & 1.369 & 1.385 & 1.384 & 1.371 & 1.116 & 1.118 & 1.112 & 1.104 & 0.886 & 0.892 & 0.882 & 0.879 & 0.722 & 0.720 & 0.709 & 0.708\\
       NTransformer  & 1.099 & 1.148 & 1.166 & 1.176 & 0.896 & 0.927 & 0.936 & 0.947 & 0.678 & 0.703 & 0.715 & 0.722 & 0.553 & 0.567 & 0.574 & 0.581\\
       Crossformer  & 1.177 & 1.278 & 1.286 & 1.293 & 0.959 & 1.032 & 1.033 & 1.041 & 0.750 & 0.825 & 0.822 & 0.826 & 0.611 & 0.666 & 0.661 & 0.665\\
       TimesNet     & 1.076 & 1.120 & 1.152 & 1.157 & 0.877 & 0.910 & 0.925 & 0.932 & 0.665 & 0.691 & 0.706 & 0.710 & 0.542 & 0.558 & 0.567 & 0.572\\
       PatchTST     &  1.041 & 1.122 & \underline{1.117} & \underline{1.087} & 0.848 & 0.904 & \underline{0.897} & \underline{0.877} & 0.640 & 0.685 & \underline{0.678} & \underline{0.660} & 0.522 & 0.551 & \underline{0.545} & \underline{0.533}\\
       iTransformer & \underline{1.030} & \underline{1.096} & 1.131 & 1.142 &\underline{0.839} &\underline{0.885}&0.909&0.920&\underline{0.632}&\underline{0.665}&0.685&0.696&\underline{0.515}&\underline{0.537}&0.551&0.561\\
       STHD   & \textbf{0.745} & \textbf{0.795} & \textbf{0.824} & \textbf{0.833} & \textbf{0.551} & \textbf{0.580} & \textbf{0.598} & \textbf{0.604} & \textbf{0.448} & \textbf{0.476} & \textbf{0.498} & \textbf{0.507} & \textbf{0.331} & \textbf{0.348} & \textbf{0.361} & \textbf{0.368} \\ \hline
       {Dataset}    & \multicolumn{15}{c}{Wiki-People}\\ \hline
       & \multicolumn{4}{c|}{RMSE} & \multicolumn{4}{c|}{WRMSPE} & \multicolumn{4}{c|}{MAE} & \multicolumn{4}{c}{WAPE}\\ \hline
       Horizon     & 28 & \ \ 35 & 42 & \ \ 49 & \ \ 28 & \ \ 35 & \ \ 42 & \ \ 49  &  \ \ 28 & \ 35 & \ 42 & \ 49  &  \ \ 28 & \ \ 35 & \ \ 42 & \ 49\\ \hline
       Naive        &  1.696 & 1.741 & 1.780 & \ \ \ - & 1.539 & 1.576 & 1.609 &  \ \ \ - & 0.993 & 1.025 & 1.056 &  \ \ \ - & 0.901 & 0.928 & 0.954 & \ \ \ - \\
       Linear & 1.433 & 1.492 & 1.552 & 1.601 & 1.300 & 1.351 & 1.402 & 1.443 & 0.805 & 0.840 & 0.878 & 0.906 & 0.731 & 0.760 & 0.793 & 0.817 \\
       DLinear & 1.373 & 1.434 & 1.500 & 1.569 & 1.246 & 1.298 & 1.355 & 1.414 & 0.767 & 0.805 & 0.840 & 0.880 & 0.696 & 0.729 & 0.759 & 0.793 \\
       Transformer & 1.920 & 1.919 & 1.922 & 1.928 & 1.742 & 1.738 & 1.736 & 1.737 & 1.193 & 1.197 & 1.191 & 1.196 & 1.082 & 1.083 & 1.076 & 1.078 \\
       NTransformer & 1.455 & 1.497 & 1.517 & 1.551 & 1.320 & 1.355 & 1.371 & 1.398 & 0.833 & 0.862 & 0.876 & 0.899 & 0.756 & 0.781 & 0.791 & 0.810 \\
       Crossformer & 1.669 & 1.698 & 1.719 & 1.745 & 1.514 & 1.537 & 1.553 & 1.573 & 0.929 & 0.947 & 0.972 & 0.995 & 0.843 & 0.858 & 0.878 & 0.897 \\
       TimesNet & 1.440 & 1.471 & 1.505 & 1.539 & 1.307 & 1.332 & 1.360 & 1.387 & 0.822 & 0.842 & 0.865 & 0.886 & 0.746 & 0.762 & 0.781 & 0.798 \\
       PatchTST & 1.387 & 1.430 & 1.462 & 1.497 & 1.259 & 1.294 & 1.321 & 1.349 & 0.783 & 0.810 & 0.830 & 0.852 & 0.710 & 0.733 & 0.750 & 0.768 \\
       iTransformer & \underline{1.361} & \underline{1.408} & \underline{1.451} & \underline{1.487} & \underline{1.235} & \underline{1.275} & \underline{1.311} & \underline{1.340} & \underline{0.758} & \underline{0.789}& \underline{0.816} & \underline{0.837} & \underline{0.688} & \underline{0.714} & \underline{0.737} & \underline{0.754} \\
       sthd &\textbf{1.188} & \textbf{1.231} & \textbf{1.272} & \textbf{1.307} & \textbf{1.141} & \textbf{1.180} & \textbf{1.216} & \textbf{1.246} & \textbf{0.658} & \textbf{0.684} & \textbf{0.708} & \textbf{0.732} & \textbf{0.632} & \textbf{0.655} & \textbf{0.676} & \textbf{0.698} \\ \hline
       {Dataset}    & \multicolumn{15}{c}{Traffic}\\ \hline
       & \multicolumn{4}{c|}{RMSE} & \multicolumn{4}{c|}{WRMSPE} & \multicolumn{4}{c|}{MAE} & \multicolumn{4}{c}{WAPE}\\ \hline
       Horizon      & 96 &  \ 192& 336 & \ 720 & \ \ 96 & \ 192& \ 336 & \ 720 & \ \ 96 & \ 192& \ 336 & 720& \ \ 96 & \ 192& \ 336 & \ 720\\ \hline
       Naive        & 1.199 & 0.968 & 0.807 & \ \ \ - & 1.500 & 1.209 & 1.005 & \ \ \ - & 0.562 & 0.417 & 0.311 & \ \ \ - & 0.704 & 0.521 & 0.387 & \ \ \ -\\
       Linear      & 0.650 & 0.669 & 0.673 & 0.698 & 0.813 & 0.835 & 0.839 & 0.867 & \underline{0.301} & 0.320 & 0.320 & 0.341 & \underline{0.377} & 0.400 & 0.398 & 0.424\\
        DLinear     & 0.651 & 0.670 & 0.676 & 0.694 & 0.815 & 0.836 & 0.841 & 0.862 & \underline{0.304} & 0.322 & \underline{0.324} & 0.335 & 0.380 & 0.402 & 0.404 & 0.417\\
       Transformer   &  0.875 & 0.889 & 0.862 & 0.855 & 1.096 & 1.110 & 1.074 & 1.062 & 0.431 & 0.446 & 0.421 & 0.411 & 0.539 & 0.556 & 0.524 & 0.511\\
       NTransformer  &  0.809 & 0.813 & 0.838 & 0.849 & 1.012 & 1.015 & 1.044 & 1.054 & 0.381 & 0.382 & 0.410 & 0.412 & 0.477 & 0.477 & 0.510 & 0.512\\
       Crossformer  & 0.826 & 0.864 & 0.869 &  \ \ \ - & 1.033 & 1.079 & 1.082 &  \ \ \ - & 0.400 & 0.432 & 0.431 & \ \ \ - & 0.501 & 0.540 & 0.536 &  \ \ \ - \\
       TimesNet     &  0.776 & 0.786 & 0.794 & 0.808 & 0.971 & 0.981 & 0.989 & 1.004 & 0.329 & 0.338 & 0.346 & 0.346 & 0.411 & 0.422 & 0.431 & 0.430\\
       PatchTST     & \underline{0.647} & \underline{0.657} & 0.668 & \underline{0.649} & \underline{0.809} & \underline{0.820} & 0.832 & \underline{0.854} & 0.313 & \underline{0.316} & 0.324 & \textbf{0.291} & 0.392 & \underline{0.394} & 0.404 & \textbf{0.383}\\
       iTransformer &0.723&0.686&\underline{0.666}& 0.722&0.904&0.856&\underline{0.830}&0.897&0.370&0.323&0.298&0.351&0.463&0.403&\underline{0.371}&0.436\\
       STHD        & \textbf{0.581} & \textbf{0.602} & \textbf{0.608} & \textbf{0.648} & \textbf{0.723} & \textbf{0.747} & \textbf{0.753} & \textbf{0.801} & \textbf{0.281} & \textbf{0.292} & \textbf{0.297} & \underline{0.319} & \textbf{0.350} & \textbf{0.363} & \textbf{0.368} & \underline{0.394}\\ \hline
          
      \hline
    \end{tabular}
    \begin{tablenotes}
        \small
        \item `-' indicates the results couldn't be obtained due to the methods setting or GPU memory limitation. For the Naive method, if the input length is less than the output length, then it cannot be predicted. For GPU memory issues, please refer to Memory Discussion in Section~\ref{sec:result_comparison}.
        \end{tablenotes}
    \end{threeparttable}
\end{table*}

\textit{Crime-Chicago}~\footnote{\url{https://data.cityofchicago.org/Public-Safety/Crimes-2001-to-Present/ijzp-q8t2}} extracted from the Chicago Police Department's Citizen Law Enforcement Analysis and Reporting System, reflects reported incidents of crime that occurred in the City of Chicago. We reserve the top 14 frequent crime types, aggregate other types of crime as one type across 77 communities, and process it as the monthly time series.  
\textit{Wiki-People}~\footnote{\url{https://www.kaggle.com/competitions/web-traffic-time-series-forecasting/}} records Wikipedia web traffic and was published on Kaggle competition. It is daily time series data. When processing, first we filter out the 27,786 series with missing data and 9,194 series not published in Wikipedia agent.
Then we filter articles that do not have series on all access, which results in 16,117 articles (93,020 series). 
Finally, we selected the `people' topic, which resulted in 1,013 articles (6,107 series). 
\textit{Traffic}~\footnote{\url{https://pems.dot.ca.gov/}} records the hourly time series of road occupancy rates measured by different sensors on San Francisco Bay area freeways. 

\subsection{Baselines and Experimental Settings}
\label{ssct-settings}
\paragraph{Baselines}
We compare our STHD model with the following representative and state-of-the-art algorithms, including 1) channel-independent models: Linear, DLinear, Transformer, Non-stationary Transformer, PatchTST, TimesNet; 2) channel-dependent models: Crossformer, iTransformer; and 3) the Na\"ive method, the most basic forecasting baseline that the forecasting value equal to the past value. All the methods apart from Na\"ive are implemented on TSlib~\footnote{\url{https://github.com/thuml/Time-Series-Library}}. 
\paragraph{Parameter settings}
All of models follow the same horizon length $\tau \in \{6, 12, 18, 24\}$ for Crime-Chicago, $\tau \in \{28, 35, 42, 49\}$ for Wiki-People, and $\tau \in \{96, 192, 336, 720\}$ for Traffic. We use long-term horizons on Traffic dataset to facilitate comparisons with existing studies, thereby simplifying the process of contrasting our findings with established research~\cite{nie_time_2023}.
For the Traffic dataset, we follow most of the parameter settings as PatchTST, leaving the test batch size larger, at the same time, ensuring all the test samples are compared.
Then we use grid-search to tune parameters, specifically, embedding size for fully connected layers $\in \{64,128,256,384\}$, embedding size for model $\in \{64,128,256,384\}$, learning rate $\in \{0.1,0.01,0.001,0.0001\}$, encoder layer, decoder layer and attention head $\in \{1,2,4,6\}$ if used.
Patch length $L=12$ for Crime-Chicago, $L=14$ for Wiki-People, and $L=24$ for Traffic and stride of patches $s=6$ for Crime-Chicago, $s=7$ for Wiki-people, and $s=12$ for Traffic.
More details about baselines are listed in the Appendix.
\paragraph{Evaluation Metrics}
Referring to ~\cite{hewamalage_forecast_2023}, 
we select metrics covering different aspects: 
rooted mean squared measure without scaling: Rooted Mean Squared Error(RMSE), rooted mean squared measure scaling with a sum over test set: Weighted Root Mean Squared Percentage Error(WRMSPE), absolute measure without scaling: Mean Absolute Error(MAE), absolute measure scaling with a sum over test set: Weighted Absolute Percentage Error(WAPE). All metrics are the measurement of errors, which means they are the smaller the better.

\subsection{Results Comparison}
\label{sec:result_comparison}

Table~\ref{tab:main_results} presents the performance of all comparison methods on the three real-world datasets, where several interesting findings are observed as follows.
1) As the simplest baseline - Na\"ive, where the forecasting value equals its closest value. It is competitive with basic deep learning model: Linear and Transformer;
2) Linear-based methods, including Linear and DLinear, 
perform better than the traditional transformer-based models but worse than the patch-based transformer models in general;
3) Transformer-based channel-independent models, including Transformer, Non-stationary Transformer, TimesNet, and PatchTST demonstrate quite different performance on three datasets. PatchTST consistently performs better on three datasets, verifying the effectiveness of patches for capturing temporal dependencies on high-dimensional MTS data;
4) The transformer-based channel-dependent baselines, iTransformer performs better than Crossformer, even overall better than channel-independent models on Wiki-People. Crossformer shows a quite stable performance that can be ranked at the middle of all baselines on high-dimensional MTS data. However, during the evaluation process, it is very time-consuming to run both, making it hard to scale well on high-dimension MTS data due to its modeling of full relations of all channels;
5) Finally, our STHD earns an overall best performance on all three high-dimensional datasets, with improvements of 30.02\%, 10.20\%, and 5.43\% on Crime-Chicago, Wiki-People, and Traffic respectively. 
\paragraph{Memory Discussion.} Training with a single A100 graphic card, CUDA out-of-memory error occurs when we use Crossformer at 720 horizons on the Traffic dataset. Since Traffic has a large $S$, samples of each batch are allocated $M$ times more than using ReIndex, which costs more parameters.

\subsection{Related Series}
\begin{figure}[htp]
    \begin{subfigure}{.19\textwidth}
        \centering
        \includegraphics[width=\linewidth]{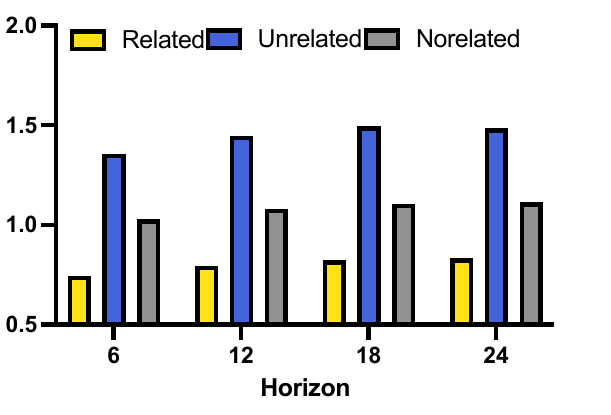}
        \caption{RMSE}
        \label{fig:k_rmse}
    \end{subfigure}
    \hspace{0.3cm}
    \begin{subfigure}{.19\textwidth}
        \centering
        \includegraphics[width=\linewidth]{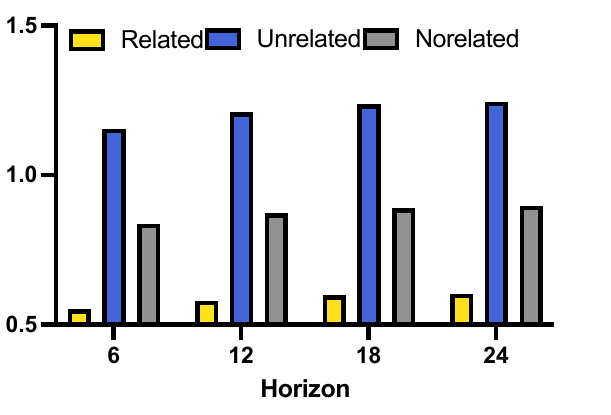}
        \caption{WRMSPE}
        \label{fig:k_wrmspe}
    \end{subfigure}
    \begin{subfigure}{.19\textwidth}
        \centering
        \includegraphics[width=\linewidth]{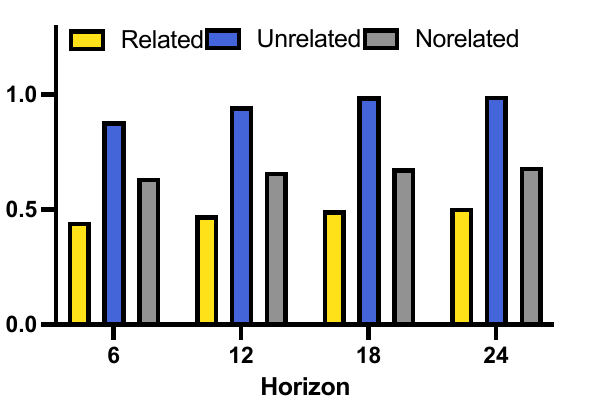}
        \caption{MAE}
        \label{fig:k_mae}
    \end{subfigure}
    \hspace{0.3cm}
    \begin{subfigure}{.19\textwidth}
        \centering
        \includegraphics[width=\linewidth]{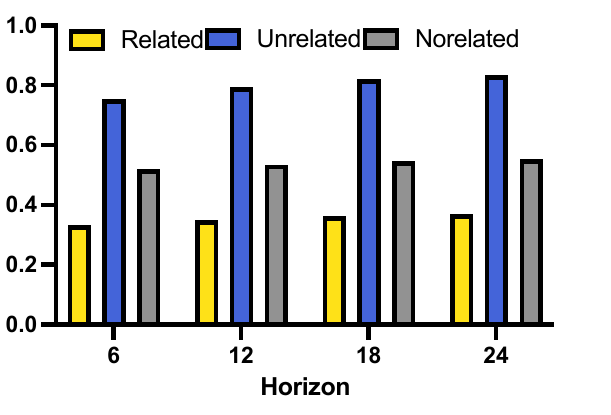}
        \caption{WAPE}
        \label{fig:k_wape}
    \end{subfigure}
    \caption{Effect of Related Series on Crime-Chicago. Yellow, blue and grey bars represent the target series with $K$ related series, with $K$ unrelated series, and without auxiliary series.}
    \label{fig:relate}
    \Description{}
\end{figure}

To better demonstrate the impact of the related series, we made the following comparison: a) target series with top $K$ related series as the input; b) target series with top $K$ unrelated series as input to our method STHD; c) target series without any series. Figure~\ref{fig:relate} shows the result on Crime-Chicago. Yellow, blue, and grey bars represent experimental results from a), b), and c) separately.
With related series, yellow bars are overall twice as good as the blue bars across all measures. 
Without related series, grey bars are overall better than blue bars, which demonstrates the importance of reducing the introduction of noise from unrelated series.
To sum up, the related series significantly benefits the task, demonstrated by its improved performance on different measures.

\subsection{Efficiency of DeepGraph} 
To verify the efficiency of DeepGraph, we compare the run time of
\noindent 
\begin{minipage}{0.24\textwidth} 
    \includegraphics[width=\textwidth]{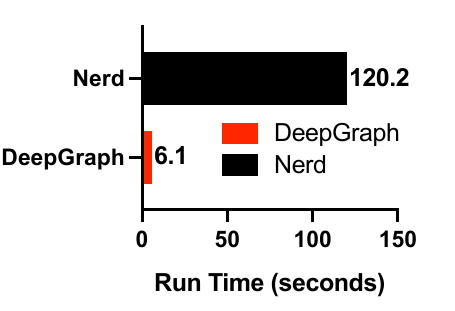} 
    \captionof{figure}{Comparison of the Run Time for Correlation Computing.}
    \label{fig:runtime}
\end{minipage}
\hfill
\begin{minipage}{0.2\textwidth}
    computing correlations with the DeepGraph graph structure and computing without DeepGraph (i.e., We use the correlation method embedded in Nerd instead of DeepGraph). We compare them to the Wiki-People dataset considering its largest number of dimensions compared with the other two datasets.  \ 
\end{minipage}
\ \ \ \ \ Running time is shown in Figure~\ref{fig:runtime}, we can observe that Nerd in black bars is 120.2 seconds, while DeepGraph in red bars are 6.1 seconds. In this way, we achieve a 20 times speedup compared with a na\"ive pair-wise computation.
\subsection{Correlation Sparsity}
\begin{figure}[htp]
    \begin{subfigure}{.155\textwidth}
        \centering
        \includegraphics[width=\linewidth]{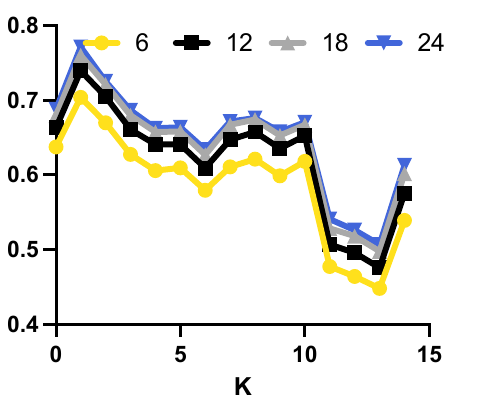}
        \caption{Crime-Chicago}
        \label{fig:crime_k}
    \end{subfigure}
    \begin{subfigure}{.155\textwidth}
        \centering
        \includegraphics[width=\linewidth]{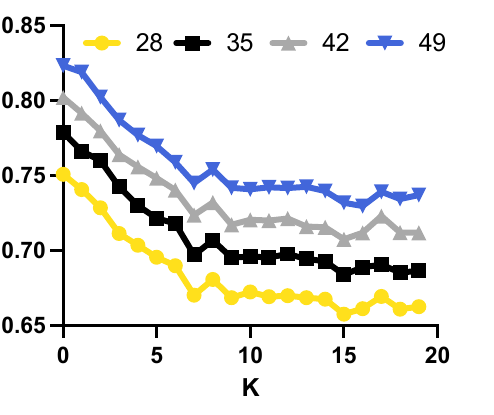}
        \caption{Wiki-People}
        \label{fig:wiki_k}
    \end{subfigure}
    \begin{subfigure}{.155\textwidth}
        \centering
        \includegraphics[width=\linewidth]{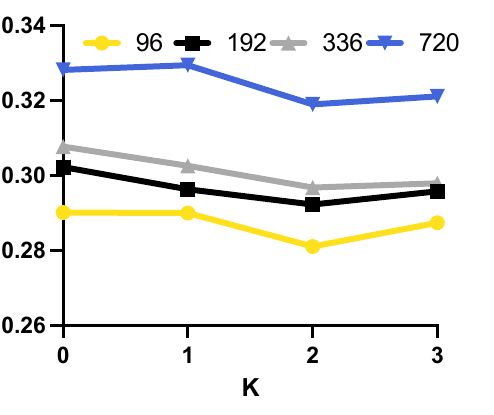}
        \caption{Traffic}
        \label{fig:traffic_k}
    \end{subfigure}
    \caption{MAE measure with the change of $K$ on three datasets. The x-axis is the value of $K$, the y-axis is MAE, and the line in yellow, black, grey, and blue represents the horizon in 6,12,18,24 separately.}
    \label{fig:k}
    \Description{}
\end{figure}

Figure~\ref{fig:k} shows the experimental results for the change of hyper-parameter $K$ on Crime-Chicago, Wiki-People, and Traffic, respectively. 
A similar trend with the change of $K$ on all the datasets is observed. With the increase of $K$, the performance first obviously drop, and then slightly rise. The possible reason is that at the very beginning, the increased $K$ means the model can benefit from real related series, while when $K$ becomes too large, the series becomes less relevant which will introduce noise to the model. 
\subsection{ReIndex}
Note that ReIndex is designed to alleviate the memory issue in computing the attention map and it makes the training stage scalable to larger batch sizes, as shown in Figure \ref{fig:reindex} and Appendix. However, we find that apart from this, ReIndex is also able to improve the model's effectiveness as shown in Table~\ref{tab:reindex}.
In this table, the following comparison is conducted on the Crime-Chicago dataset: take the target series as the input, and use STHD with ReIndex $STHD_{wt}$ and without ReIndex $STHD_{wto}$ to forecast.
The results show that
$STHD_{wt}$ outperforms $STHD_{wto}$ across all measures across all horizons. 
\begin{table}[ht]
    \caption{Experimental results for ReIndex. $STHD_{wt}$ and $STHD_{wto}$ represent the extensive experiment that takes target series as input for STHD with ReIndex and without ReIndex respectively. Best results are shown with \underline{underline}.}
    \centering
    \label{tab:reindex}
    \renewcommand{\arraystretch}{0.6}
    \small
    \begin{tabular}{p{1.1cm}|p{0.43cm}|p{0.43cm}|p{0.43cm}|p{0.43cm}|p{0.43cm}|p{0.43cm}|p{0.43cm}|p{0.43cm}|p{0.43cm}} \hline
    & \multicolumn{3}{c|}{RMSE} & \multicolumn{3}{c|}{WRMSPE} & \multicolumn{3}{c}{WAPE} \\ \hline
      Horizon      & \ \ 6 & \ \ 12 & \ 18 & \ \ 6 & \ \ 12 & \ 18 & \ \ 6 & \ 12 & \ \ 18 \\ \hline
        $STHD_{wt}$ & \underline{1.028} & \underline{1.081} & \underline{1.107} & \underline{0.838} & \underline{0.872} & \underline{0.889} &
        \underline{0.519} & 
        \underline{0.535} &
        \underline{0.545}\\
        $STHD_{wto}$ & 1.041 & 1.122 & 1.117  & 0.849 & 0.904 & 0.898 & 0.522 & 0.551 & 0.545 \\
        \hline
    \end{tabular}
\end{table}

\section{Conclusion}
This paper proposed a scalable Transformer framework - STHD, designed specifically for tackling the intricate challenges of high-dimensional Multivariate Time Series (MTS) forecasting.
Through the innovative incorporation of sparse correlation matrix, STHD significantly enhances forecasting performance by efficiently filtering out unrelated series.
Furthermore, the introduction of the ReIndex strategy emerges as a dual-purpose solution, mitigating the constraints of parameter ranges in training transformer-based methods on high-dimensional MTS data, while at the same time enriching the diversity of shuffled training samples.
Experimental results underscore STHD's forecasting performance in high-dimensional MTS data. In essence, STHD model focuses on the correlation of high-dimensional data with accurate forecasts.
Further, we will continue to refine and expand upon the STHD model, with modalities and will further explore the co-evaluation of high-dimensional MTS data.

\section*{Appendix}
\appendix
\section{Data Analysis and Processing}
\begin{figure}[ht]
    \centering
    \includegraphics[width=0.4\linewidth]{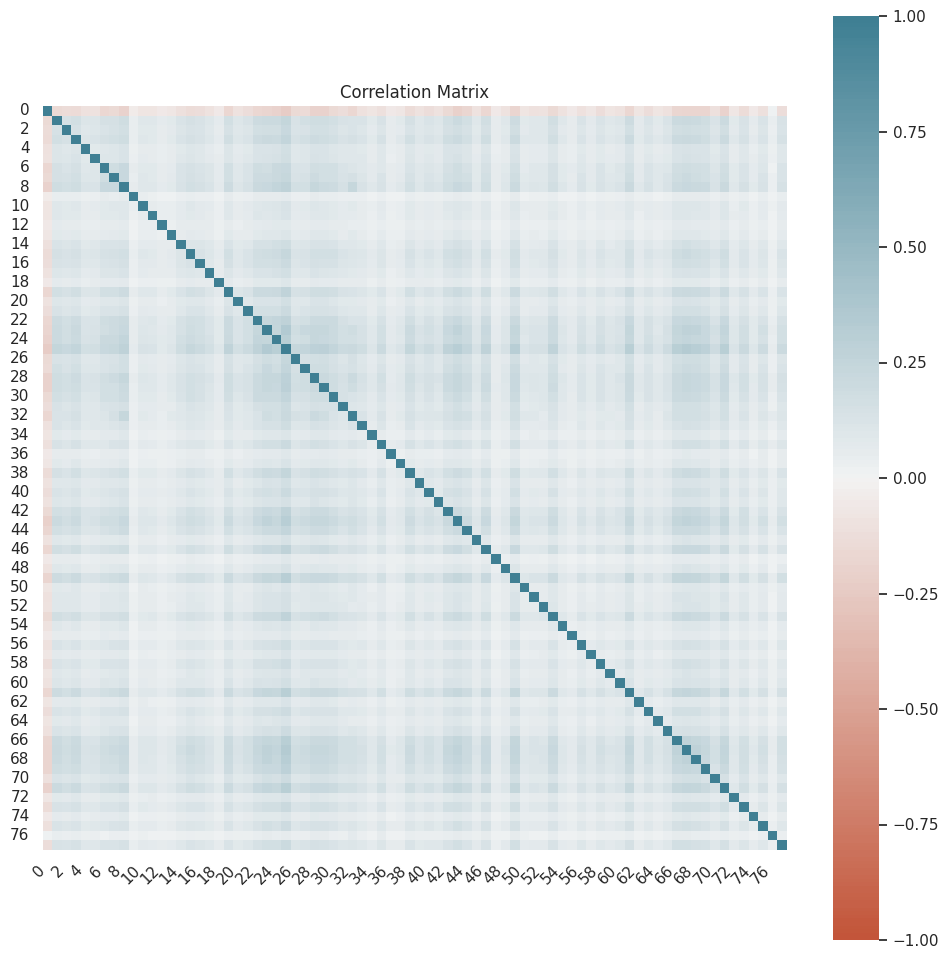}
    \caption{The correlation of communities on Crime-Chicago. The color scale on the right indicates the strength and direction of the correlation coefficient.}
    \label{fig:crime-rel}
    \Description{}
\end{figure}

Before this work, none of the time series works used the Crime-Chicago dataset for MTS forecasting, so we conducted data analysis for Crime-Chicago. Figure~\ref{fig:crime-rel} shows the correlation of different communities, the x-axis, and y-axis represent the community area. To draw this figure, we add up all crime types of the series from the same community. The color scale on the right indicates the strength and direction of the correlation coefficient:
Dark blue represents a correlation coefficient close to 1, indicating a strong positive correlation; Lighter blue and white represent correlation coefficients closer to 0, indicating little or no linear correlation; The red shades (which are not prominently visible in this matrix) would indicate a negative correlation, with dark red being close to -1. Through the figure, we can clearly see different color distributions, which illustrates why STHD uses the correlation method. For more details about analyzing and pre-processing, please refer to Colab~\footnote{\url{https://colab.research.google.com/drive/1u9utf8ZadWTWo73s_l4Uh6dETLvleNts?usp=sharing}}.

\section{Theoretical Analysis}
\paragraph{Time.} Relation Matrix Sparsity module, with its $O(M^2)$ complexity of $M$ series, is the most computationally intensive part, primarily due to pairwise correlation computations. Because of that, we leverage parallel computing and efficient data structures to mitigate the computational load. 
The ReIndex module, while crucial for data optimization, earns the same complexity as other works. 
The Transformer-based model's primary complexity stems from its attention mechanism. Unlike PatchTST, which focuses solely on time dependency with a complexity of $O(P^2)$ with $P$ patches within a subseries, Adapted Transformer's complexity is primarily dependent on the attention: $O(((1+K)SP)^2)$ for $S$ subseries of each series. However, previous Sparsity and Reindex steps effectively reduce the input size where $K<20, (1+K) << M$, mitigating the quadratic dependency. 
Improvements in accuracy for high-dimensional datasets also counterbalance this.

\paragraph{Memory.} Relation Matrix Sparsity uses $O(MK)$ where $K << M$ is the most significant correlation. For $M$ time series, the naive method stores a full correlation matrix that requires $O(M^2)$. ReIndex reduces samples from $b \times M \times (1+K)$ to $b \times (1+K)$. Adapted Transformer: needs extra memory than other transformers for attention $O((KS)^2)$ and related layers.

\section{Experimental Setting} 
All DNNs models including pre-test, ablation study on STHD, and baselines are implemented in PyTorch and trained on a single NVIDIA A100 GPU. 
The parameter configurations for the Crime-Chicago and Wiki-People are outlined in Table~\ref{tab:param_crime}. For Traffic, we follow the setting of PatchTST. To allow each model to converge during training, we set the training epoch to 100, and use an early stop setting, which means if the delta of the validate set loss is less than e-7, the training will stop and begin to test. 

\begin{table}[h]
    \caption{Parameters in Crime-Chicago and Wiki-People.}
    \centering
    \label{tab:param_crime}
    \tiny 
    \begin{tabular}{c|c|c|c|c|c|c}\hline 
     {Dataset}    & \multicolumn{6}{c}{Chicago-Crime}\\ \hline
         &  $d_{ff}$& $d_{model}$ & $e_{layer}$ & $lr$ & $b$ & $n_{head}$\\\hline
Linear  & - & 128 & - & 0.01 & 128 & -\\
DLinear  & - & 384 & - & 0.0001 & 128 & -\\
Transformer& 384 & 256 & 2 & 0.001 & 32 & 4\\
NTransformer& 384 & 256 & 1 & 0.001 & 32 & 4\\
Crossformer& 384 & 256 & 2 & 0.01 & 32 & 4\\
PatchTST& 384 & 256 & 2 & 0.001 & 32 & 4\\
TimesNet& 384 & 256 & 2 & 0.001 & 32 & 4\\
iTransformer& 384 & 256 & 2 & 0.001 & 32 & 4\\
STHD& 384 & 256 & 2 & 0.001 & 128 & 4\\ \hline
    {Dataset}    & \multicolumn{6}{c}{Wiki-People}\\ \hline
    Linear  & - & 128 & - & 0.01 & 16 & -\\
    DLinear  & - & 128 & - & 0.01 & 32 & -\\
    Transformer& 128 & 128 & 1 & 0.001 & 32 & 4\\
    NTransformer& 128 & 128 & 1 & 0.001 & 32 & 4\\
    Crossformer& 128 & 128 & 2 & 0.001 & 32 & 4\\
    PatchTST&128 & 128 & 1 & 0.001 & 32 & 4\\
    TimesNet& 128 & 128 & 2 & 0.001 & 32 & 4\\
    iTransformer&128 & 128 & 1 & 0.001 & 32 & 4\\
    STHD& 128 & 128 & 1 & 0.001 & 128 & 4\\ \hline
    \end{tabular}
\end{table}

\section{Performance on Low-dim Datasets}
While the STHD is designed with high-dimensional MTS forecasting, theoretically, it can be applied to low-dimensional series. Some further experiments on lower-dimensional data have been done on Weather. Results are listed in Table~\ref{tab:results_on_lowdim}.

\begin{table}[ht]
    \caption{Experimental results on low-dimensional dataset. The best results are in \textbf{bold} and the second best \underline{underlined}.}
    \label{tab:results_on_lowdim}
    \centering 
    \renewcommand{\arraystretch}{0.95}
    \tiny
    \begin{tabular}{c|c|c|c|c|c|c|c|c}\hline 
       & \multicolumn{4}{c|}{MAE} & \multicolumn{4}{c}{RMSE} \\ \hline
       Horizon     & 96 & 192 & 336 & 720 &  96 & 192 & 336 & 720  \\ \hline
       DLinear & 0.243 & 0.422 & 0.281 & 0.468 & 0.320 & 0.515 & 0.372 & 0.574 \\
       Crossformer & 0.214 & \underline{0.384} & 0.265 & \underline{0.442} & 0.335 & 0.517 & 0.416 & 0.605 \\
       PatchTST & \underline{0.205} & 0.392 & \underline{0.245} & 0.444 & \underline{0.284} & 0.498 & 0.335 & 0.567 \\
       TimesNet & 0.274 & 0.399 & 0.287 & 0.505 & 0.287 & \textbf{0.419} & \textbf{0.333} & \textbf{0.482} \\
       iTransformer & 0.211 & 0.401 & 0.251 & 0.453 & 0.290 & 0.506 & 0.338 & 0.571 \\
       STHD & \textbf{0.187} & \textbf{0.316} & \textbf{0.233} & \textbf{0.388} & \textbf{0.276} & \underline{0.452} & \underline{0.331} & \underline{0.530} \\ \hline
    \end{tabular}
\end{table}
We have the following findings: 1) channel-dependent model Crossformer performs well when the horizon equals 192 and 720 on MAE, which may contribute to less noise introduced in low-dimensional datasets;
2) patch-based method PatchTST and multi-frequency-based method TimesNets show their ability to capture long-term dependencies;
3) by sampling correlated series ($K=1$), STHD still works on low-dimensional datasets.

\bibliographystyle{ACM-Reference-Format}
\bibliography{myref}

\end{document}